\documentclass[sigconf]{acmart}

\usepackage{multirow}
\usepackage{makecell}

\AtBeginDocument{%
  }

\setcopyright{none}

\renewcommand\footnotetextcopyrightpermission[1]{}

\begin{document}
\begin{sloppypar}

\title{DLGAN : Time Series Synthesis Based on Dual-Layer Generative Adversarial Networks}
% \title{The Name of the Title Is Hope}


% 匿名作者信息
\author{Xuan Hou}
 \affiliation{
   \institution{Shandong University}
   \city{Qingdao}
   \country{China}
 }
    \email{houxuan@mail.sdu.edu.cn}
 
\author{Shuhan Liu}
 \affiliation{
   \institution{Shandong University}
      \city{Qingdao}
   \country{China}
 }
    \email{shuhanliu@mail.sdu.edu.cn}
 
 \author{Zhaohui Peng}
 \affiliation{
   \institution{Shandong University}
      \city{Qingdao}
   \country{China}
 }
  \email{pzh@sdu.edu.cn}
 \authornote{Corresponding author.}
 
 \author{Yaohui Chu}
 \affiliation{
   \institution{Shandong University}
      \city{Qingdao}
   \country{China}
 }
     \email{cyh0206@mail.sdu.edu.cn}
     
 \author{Yue Zhang}
 \affiliation{
   \institution{Shandong University} 
      \city{Qingdao}
   \country{China}
   }
   \email{zhangyue_zz@mail.sdu.edu.cn}
   
 \author{Yining Wang}
 \affiliation{
   \institution{Shandong University}
      \city{Qingdao}
   \country{China}
 }
     \email{wangyning@mail.sdu.edu.cn}
     
% \affiliation{%

% \affiliation{%


% 右上角页眉
\renewcommand{\shortauthors}{Hou X., Liu S. et al.}


\begin{abstract}
  Time series synthesis is an effective approach to ensuring the secure circulation of time series data. Existing time series synthesis methods typically perform temporal modeling based on random sequences to generate target sequences, which often struggle to ensure the temporal dependencies in the generated time series. Additionally, directly modeling temporal features on random sequences makes it challenging to accurately capture the feature information of the original time series. To address the above issues, we propose a simple but effective generative model \textbf{D}ual-\textbf{L}ayer \textbf{G}enerative \textbf{A}dversarial \textbf{N}etworks, named \textbf{DLGAN}. The model decomposes the time series generation process into two stages: sequence feature extraction and sequence reconstruction. First, these two stages form a complete time series autoencoder, enabling supervised learning on the original time series to ensure that the reconstruction process can restore the temporal dependencies of the sequence. Second, a Generative Adversarial Network (GAN) is used to generate synthetic feature vectors that align with the real-time sequence feature vectors, ensuring that the generator can capture the temporal features from real time series. Extensive experiments on four public datasets demonstrate the superiority of this model across various evaluation metrics.
\end{abstract}






\maketitle

\section{Introduction}

Traditional manufacturing enterprises accumulate a large amount of industrial time series data in their daily production and operational activities, but often lack the necessary tools for analysis and utilization. To fully exploit the potential of this data, collaboration with technology companies or data analysis firms is often necessary. An ideal collaboration mode involves traditional manufacturing enterprises providing an authentic and effective industrial time series dataset, and the collaborating partner utilizes this data set to develop the corresponding model algorithms and provide feedback. However, due to concerns about protecting trade secrets and user privacy, direct sharing of authentic data with the collaborating partners is typically not permitted \cite{28,29,30,31}. Facing this contradiction, synthesizing a high-quality dataset with equivalent utility is a preferable solution.

% 近年来，生成模型得到极大发展，主要包括基于生成对抗网络，基于变分自动编码器以及基于扩散模型的生成模型，尽管它们的主要影响力在计算机视觉领域用于生成真实图像，但是由于其在视觉领域的成功，研究人员开始将其应用到时序数据合成领域。这些方法试图学习时间序列固有的特性和时间依赖性，并从随机序列中生成具有相同效用以及统计特性的新时间序列。
In recent years, generative models have undergone significant development, primarily including those based on Generative Adversarial Networks (GANs)\cite{3}, Variational Autoencoders (VAEs)\cite{41}, and Diffusion Models\cite{42}. Although their impact has primarily been in the field of computer vision \cite{20,21,22,32,33} for generating realistic images, researchers have begun to apply them to the field of time series synthesis due to their success in the field of vision. These methods attempt to learn the inherent features and temporal dependencies of time series, generating new time series with similar utility and statistical properties from random sequences.

% 然而，现有方法主要存在以下两个问题。首先，这些方法往往是从特定分布采样的随机序列出发，使用生成模块来生成目标序列。然而，时间序列的生成更应该是一个确定的连续性过程，即使生成模块中采用了一些连续的时序建模工具，如RNN模型及其变种LSTM、GRU等，他们的主要作用是捕获输入序列数据的模式和规律，但由于输入数据是离散的随机序列，其本身并没有任何显著的时序模式和规律，导致这些时序建模工具的表现不佳，无法有效保证生成序列的时序依赖性。其次，时间序列往往具有较强的时序特征，而这些模型从随机序列到目标序列的生成过程往往是无监督的，并且同样由于输入数据是离散的随机序列，一些在时间序列表示以及预测方面行之有效的方法也难以直接应用到时间序列生成过程中，导致模型无法充分学习到原始时间序列的特征信息。

% 现有方法往往从特定分布采样的随机序列出发，在其基础上进行连续时序建模，以生成目标时间序列，然而，这些时序建模工具，如RNN模型及其变种LSTM、GRU等，他们的主要作用是捕获输入序列数据的模式和规律，但由于输入数据是离散的随机序列，其本身并没有任何显著的时序模式和规律，导致这些时序建模工具的表现不佳，无法有效保证生成序列的时序依赖性。其次，时间序列往往具有较强的时序特征，而这些模型从随机序列到目标序列的生成过程往往是无监督的，并且同样由于输入数据是离散的随机序列，一些在时间序列表示以及预测方面行之有效的方法也难以直接应用到时间序列生成过程中，导致模型无法充分学习到原始时间序列的特征信息。

Existing methods tend to start with random sequences sampled from a particular distribution and perform continuous temporal modeling on them to generate target time series. However, temporal modeling tools such as RNNs \cite{27} and their variants LSTM\cite{4}, GRU\cite{5}, etc., primarily aim to capture patterns and regularities in input sequence data. Since the input data consists of discrete random sequences that inherently lack significant temporal patterns or regularities, these modeling tools often perform poorly, failing to effectively ensure the temporal dependencies in the generated sequences. Besides, time series often exhibit strong temporal features, such as local dependencies, periodicity, etc. \cite{34,36,37,38}. The generation process from random sequences to target sequences in these models is typically unsupervised, and also due to the fact that the input data are discrete random sequences, some of the methods that are proven to be effective in time series representation and prediction \cite{11,13,14,15,19,34} are difficult to directly apply to the time series generation process, resulting in the models not being able to fully learn the temporal feature of the original time series.

% 针对上述问题，本文提出一个简单而有效的时间序列生成模型双层生成对抗网络，叫做DLGAN。通过对时间序列合成过程进行有效分解，DLGAN将时间序列生成过程分解为序列特征提取与序列重建过程。首先，这两个过程构成一个完整的时间序列自编码器，可以在原始时序数据集上进行完整的有监督学习，一方面保证模型能够有效提取原始时间序列的时序特征信息，另一方面，特征重建过程可以保证从特征向量重建的时间序列的序列依赖性。其次，我们在上述时序自编码过程中加入两层生成对抗网络，第一层生成网络负责生成经过特征提取的时序特征向量，以此来保证生成过程可以感知到来自真实时间序列的时序特征；第二层生成网络则为时序特征重建过程加上一层无监督训练，提高特征重建过程对时序特征向量的解释重建能力。改论文的主要贡献如下：

% 由于GAN在特征向量生成的难度要远低于直接生成目标序列，因此第一层网络保证生成与时序特征提取过程提取的原始特征向量对齐的时序特征向量，而非直接生成目标序列，在保证生成过程可以感知到来自原始时间序列的时序特征的同时充分发挥生成对抗网络在特征向量生成方面的优势；第二层网络为特征重建过程加上一层无监督训练，在有监督训练保证生成序列时序依赖的同时，无监督训练提高特征重建过程对生成特征向量的解释重建能力，在一定程度上可以放松对特征向量合成质量的要求。此时，两种训练过程的目的都是根据特征向量进行时间序列的重建，因此可以有效组合来提高生成序列的质量。该论文的主要贡献如下：
In response to the above issues, we propose a simple but effective generative model Dual-Layer Generative Adversarial Networks, named DLGAN. The model decomposes the time series generation process into two stages: sequence feature extraction and sequence reconstruction. First, these two stages form a complete time series autoencoder, allowing for comprehensive supervised learning on the original time series dataset. This ensures that the model can effectively extract the temporal features from the original time series. Additionally, the feature-reconstruction process ensures the temporal dependencies of the time series reconstructed from the feature vectors. Second, we add two layers of GANs to the above time series autocoding process. The first-layer generative network is responsible for generating temporally extracted feature vectors, ensuring that the generation process can capture the temporal characteristics of real-time series. The second-layer generative network adds an unsupervised training layer to the temporal feature reconstruction process, enhancing the reconstruction process's ability to interpret and rebuild the temporal feature vectors. The main contributions of this paper are as follows.

% 提出了DLGAN，将时间序列生成过程置于基于原始时间序列的有监督学习的基础上，通过有监督学习和无监督学习的结合来保证生成时间序列的时序依赖性
% 将时间序列特征提取过程置于原始序列上，而生成对抗网络用于生成与原始特征向量对齐的特征向量，在可以有效利用现有时序特征提取方法的同时保证生成过程也能感知到来自原始时序数据的时序特征
% 四个公开数据集上的大量实验表明DLGAN在大量评价指标上优于SOTA方法
\begin{itemize}
\item The DLGAN model is proposed, which places the time series generation process on a foundation of supervised learning based on the original time series. By combining supervised and unsupervised learning, DLGAN ensures the temporal dependencies of the generated time series.
% effectively combining the iterative autoregressive generation with GANs, addressing the issue that existing methods fails to generate synthetic time series data with time series dependency.
\item The time series feature extraction process is applied to the original sequence, while the GAN is used to generate feature vectors that align with the original temporal feature vectors. This approach effectively leverages existing temporal feature extraction methods while ensuring that the generation process also captures the temporal features from the original time series data.
% The GAN-based time series synthesis methods is improved by providing GANs with temporal features of the original dataset, addressing the issue that existing methods are difficult to simultaneously analyze the temporal features of the original dataset while learning the mapping relationship from random sequences to target sequences.
\item Extensive experiments on four publicly available datasets show that the DLGAN performs better than the state-of-the-art methods on a variety of evaluation metrics.
\end{itemize}

\begin{figure*}
  \centering
  \includegraphics[width=0.9\textwidth]{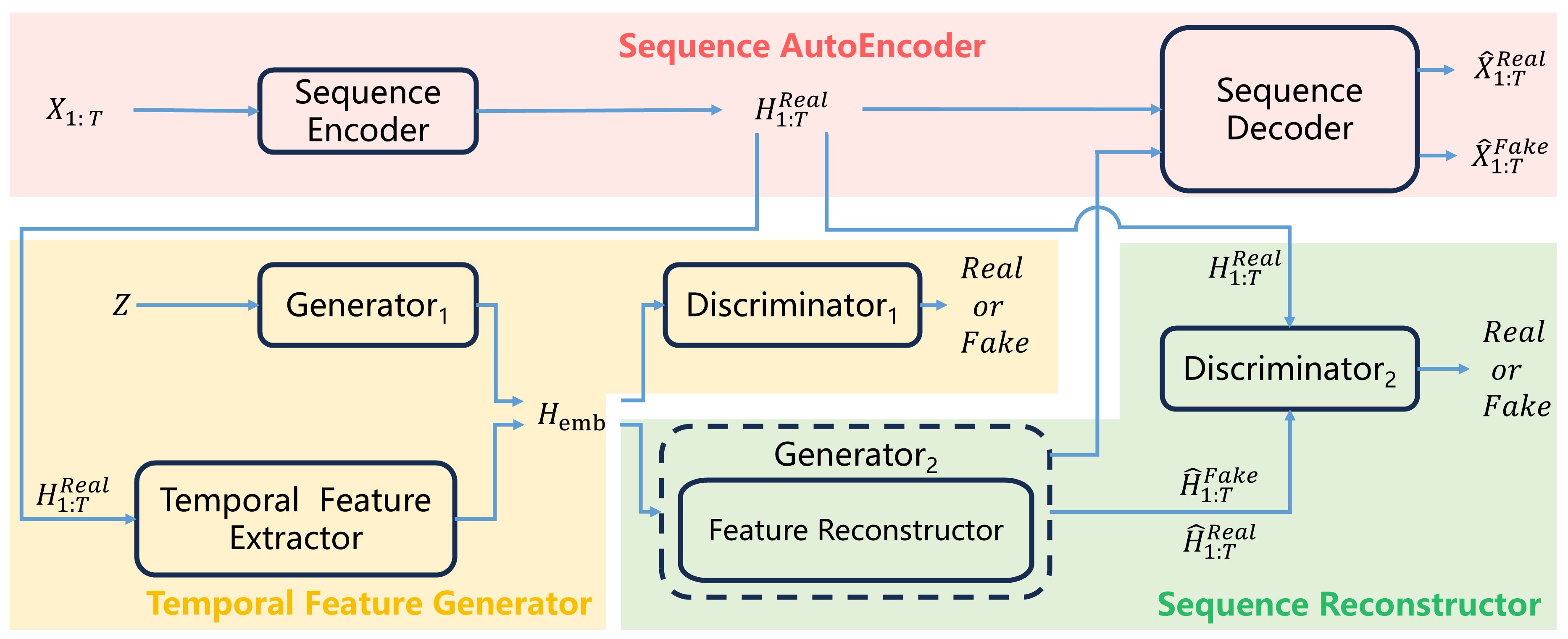}
  \caption{Overall architecture of the DLGAN}
  \label{fig:model}
\end{figure*}

\section{Related Work}

% 数据共享的需求以及对数据隐私保护的需求往往同时存在。解决问题的关键在于合成与原始数据具有相同效用的高质量合成数据集。早期的研究往往基于统计模型，尝试通过识别数据属性之间相关性来建立真实数据的近似分布，通过从该分布中进行采样来生成合成数据集。然而，数据中的复杂关系往往是不易捕获的，近似分布并不能完全还原数据的真实分布，因此，该类方法很难合成高质量的合成数据集。
The demand for data sharing and the need for data privacy protection often coexist. The critical solution to addressing this issue lies in the synthesis of high-quality synthetic datasets that have the same utility as the original data when confronted with the same task. Earlier research often relied on statistical models, attempting to establish an approximate distribution of real data by identifying correlations between attributes. Synthetic datasets were then constructed by sampling from this approximate distribution \cite{2,35}. However, capturing the complex relationships between data is challenging, and approximate distributions often struggle to fully replicate the true distribution of the original data. As a result, it is difficult to synthesise high quality synthetic datasets using such methods.

% 由于神经网络在捕获数据相关性的优越性能，已经被广泛应用到数据合成领域。在时序数据合成方面，生成对抗网络无疑是应用最为广泛的模型之一，近些年也开始出现一些基于变分自动编码器以及扩散模型的时间序列生成模型。
Due to the superior capability of neural networks in capturing data correlations, they have been widely applied in the field of data synthesis. GANs\cite{8,9,10,17,26} are undoubtedly among the most widely used models. In recent years, there has also been a rise in time series generation models based on VAEs\cite{43,44} and Diffusion Models\cite{45,46}.

% 生成对抗网络由生成器和判别器组成，生成器从随机数据出发生成目标数据，判别器通过判别原始数据和生成数据的真伪来提高模型性能。得益于生成对抗网络在关系数据生成的广泛研究，早期的时间序列生成模型使用时序建模模块对生成器和判别器进行简单重构进行时序数据生成，如，Mogren et al.使用lstm对生成器和判别器进行重构。而Yoon et al.增加一个序列回归预测过程，为生成序列提供时间动力学支持。Jeha et al.使用渐进式生成和自注意力机制来逐步生成长距离时序样本。
The GAN consists of a generator and a discriminator, the generator generates the target data from random data and the discriminator improves the model performance by discriminating the authenticity of the original and generated data. Benefiting from the extensive research on generative adversarial networks for relational data generation \cite{6,7,23,24,25}, early time series generative models used the temporal modeling module to perform a simple reconstruction of the generator and discriminator for temporal data generation, for example, Mogren et al.\cite{8} used LSTM for reconstruction of generators and discriminators. Additionally, Yoon et al.\cite{9} introduced an additional sequence regression prediction process to provide temporal dynamics support for the generated sequences. Jeha et al.\cite{10} employed progressive generation and self-attention mechanisms to gradually generate long-range temporal samples.

% 生成对抗网络由生成器和判别器组成，生成器从随机数据出发生成目标数据，判别器通过判别原始数据和生成数据的真伪来提高模型性能。得益于生成对抗网络在关系数据生成的广泛研究，早期的时间序列生成模型使用时序建模模块对生成器和判别器进行简单重构进行时序数据生成，如，Mogren et al.使用lstm对生成器和判别器进行重构。而Yoon et al.增加一个序列回归预测过程，为生成序列提供时间动力学支持。Jeha et al.使用渐进式生成和自注意力机制来逐步生成长距离时序样本。
% The GAN consists of a generator and a discriminator, the generator generates the target data from random data and the discriminator improves the model performance by discriminating the authenticity of the original and generated data. Benefiting from the extensive research on generative adversarial networks for relational data generation \cite{6,7,23,24,25}, early time series generative models used the temporal modeling module to perform a simple reconstruction of the generator and discriminator for temporal data generation, for example, Mogren et al.\cite{8} used LSTM for reconstruction of generators and discriminators. Additionally, Yoon et al.\cite{9} introduced an additional sequence regression prediction process to provide temporal dynamics support for the generated sequences. Jeha et al.\cite{10} employed progressive generation and self-attention mechanisms to gradually generate long-range temporal samples.

% 变分自动编码器由编码器和解码器组成，编码器将原始数据映射到标准正态分布，而解码器从标准正态分布中进行随机采样得到随机数据，进而生成目标数据。TimeVAE对编码器和解码器进行简单重构，使得VAE适合于时间序列生成。CR-VAE在时间序列生成过程中学习因果图，并将其纳入到数据生成过程。
The VAEs consist of an encoder and a decoder. The encoder maps the original data to a standard normal distribution, while the decoder samples randomly from this distribution to generate target data. Desai et al.\cite{43} simply reconstructs the encoder and decoder to make VAE suitable for time series generation. Li et al.\cite{44} learns causal graph during the time series generation process and incorporates it into the data generation process.

% 扩散模型通过逐步添加噪声的方式将原始数据逐步映射到标准正态分布空间，并通过学习扩散模型的逆过程将从标准正态分布空间采样得到的随机数据还原到原始数据分布空间。Diffusion-TS将季节性趋势分解技术与去噪模型结合学习时间序列的季节性趋势。MR-Diff通过季节趋势分解，从粗粒度向细粒度进行逐步建模生成目标序列。
Diffusion models gradually map original data to a standard normal distribution space by progressively adding noise, and then reverse the diffusion process to transform randomly sampled data from the standard normal distribution back to the original data distribution space. Yuan et al.\cite{45} combines seasonal trend decomposition techniques with denoising models to learn the seasonal trends of time series. Shen et al.\cite{46} performs seasonal trend decomposition and incrementally models the target sequence from coarse to fine granularity.

% 以上三类模型中，在离散的随机序列上进行连续时间序列建模的问题是普遍存在的，并不能保证生成序列的时序依赖性。并且它们往往依赖于模型特性来考虑部分时序特征，比如扩散模型天然的具备由粗粒度到细粒度的建模过程，因此该类研究开始研究多尺度时序特征信息。由于特征提取模型也难以作用到基于随机采样的输入上，生成对抗网络以及变分自动编码器则很少考虑时序特征信息。
In all three types of models, the issue of performing continuous time series modeling on discrete random sequences is very common, which fails to ensure the temporal dependencies of the generated sequences. These models often rely on specific model characteristics to consider partial temporal features. For example, diffusion models naturally possess a coarse-to-fine-grained modeling process, which has led to research into multi-scale temporal feature. However, because feature extraction modules struggle to work effectively with randomly sampled inputs, GANs and VAEs rarely consider temporal feature information.

\section{Methodology}

% 在时间序列合成任务中，我们的目标是建立一个生成模型，能够有效捕获原始时间序列的特征和依赖性，并通过随机输入来生成具有相同效用的高质量合成数据集。DLGAN主要由三个模块构成：时间序列自动编码器，将原始时间序列映射到低维的嵌入空间，捕捉其主要特征的同时提供便于后续模块学习的数据模式；时序特征生成器，使用时序特征提取模块从原始时间序列中提取时序特征向量，并生成与真实时序特征向量对齐的合成时序特征向量；时间序列重建器，结合有监督的自回归过程，从时序特征向量出发重建时间序列。我们将首先介绍模型的总体架构，然后详细描述每个组件模块。
In the task of time series synthesis, our goal is to develop a generative model that can effectively capture the features of the original time series and generating high-quality synthetic datasets with temporal dependencies from random inputs. The DLGAN mainly consists of three component modules: Sequence Autoencoder, which maps the original time series into a low-dimensional embedding space, capturing its primary features while providing  data patterns that facilitates learning in subsequent modules. Temporal Feature Generator, which uses a temporal feature extractor to extract temporal feature vectors from the original time series and generate synthetic temporal feature vectors aligned with the real temporal feature vectors. Sequence Reconstructor, which combines a supervised autoregressive process to reconstruct the time series from the temporal feature vectors. We will first present the overall architecture of the model, followed by a detailed description of each component module.

\subsection{Overall Architecture}

The overall architecture of DLGAN is illustrated in Figure \ref{fig:model}, which mainly consists of three major components: \textbf{1) Sequence Autoencoder}, highlighted by a light red background, is given the original time-series data $X_{1:T}=(X_1,\cdots,X_T )$, where $X_t=(X_t^1,X_t^2,\cdots ,X_t^M )$ represents the feature vector of the multidimensional time series at time step $t$, and $M$ is the number of attributes contained in the time series. The encoder generates a hidden sequence $H_{1:T}^{Real}=(H_1^{Real},H_2^{Real},\cdots ,H_T^{Real})$, where $H_t^{Real}=(H_t^1,H_t^2,\cdots ,H_t^N)$ represents the feature vector at time step $t$ of the hidden sequence, and $N$ is the number of attributes included in the hidden sequence. This is followed by decoding to reconstruct $\widehat{X}_{1:T}^{Real}\in R^{M\times T}$. \textbf{2) Temporal Feature Generator}, highlighted by a light yellow background, the temporal feature extractor receives the hidden sequence $H_{1:T}^{Real}$ for extracting temporal features, generating the synthesised temporal feature vector $H_{emb}^{Real}$. Meanwhile, Generator\textsubscript{1} accepts a random sequence $Z$ to generate the temporal feature vector $H_{emb}^{Fake}$, and Discriminator\textsubscript{1} discerns the authenticity of the temporal feature vectors. \textbf{3) Sequence Reconstructor}, highlighted by a light green background, Generator\textsubscript{2}, also known as the feature reconstructor, receives the temporal feature vectors to reconstruct the hidden series, generating $\widehat{H}_{1:T}^{Real}\in R^{N\times T}$ and $\widehat{H}_{1:T}^{Fake}\in R^{N\times T}$. Discriminator\textsubscript{2} is responsible for discerning the authenticity of these reconstructed hidden sequences. After training the model, random sequences are passed through Generator\textsubscript{1} and Generator\textsubscript{2} to generate $\widehat{H}_{1:T}^{Fake}$, which is then decoded to generate the synthetic time series data.

% 时间序列自动编码器
\subsection{Sequence Autoencoder}
The autoencoder consists of two major components: the sequence encoder and decoder. The encoder is responsible for mapping the original time series to a low-dimensional embedding space, while the decoder reconstructs the time series from this embedding space. The reason for choosing to map the time series to an embedding space rather than directly using the original data for subsequent training lies in two aspects. On the one hand, time series data often contain considerable noise. Mapping time series to a low-dimensional embedding space makes it easier to capture the intrinsic characteristics of the time series, thereby improving the model's generalization ability. On the other hand, through joint training, the encoder can also learn a more favorable representation of the time series for the subsequent modules' training. The encoding and decoding process can be represented as follows:

\begin{equation}
  H_{1:T}^{Real}=encoder(X_{1:T} ),
\end{equation}
\begin{equation}
  \widehat{X}_{1:T}^{Real}=decoder(H_{1:T}^{Real} ),
\end{equation}

In this case, the encoder and decoder can be implemented using any time series modeling module. Here, we choose to implement them using a deep GRU (Gated Recurrent Unit) network.

\begin{figure*}[htbp]
  \centering
  \includegraphics[width=0.8\textwidth]{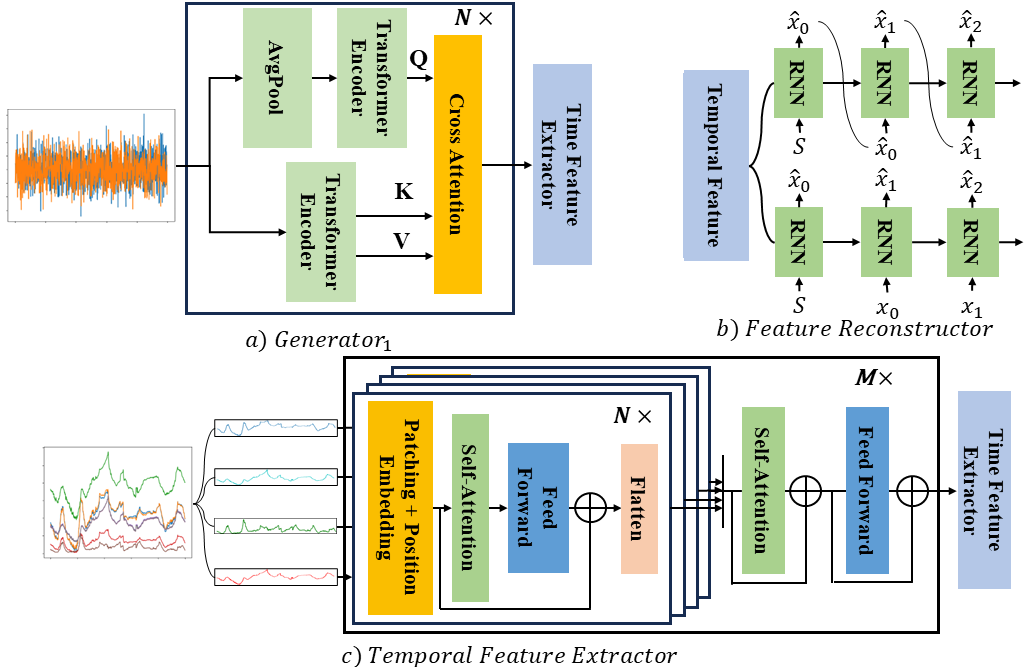}
  \caption{Module Detailed Architecture Diagram}
  \label{fig:extractor}
\end{figure*}

\subsection{Temporal Feature Generator}
Temporal Feature Generator comprises three main components: the Temporal Feature Extractor, Generator\textsubscript{1}, and Discriminator\textsubscript{1}. The critical design lies in the Temporal feature extractor's effective utilization of component analysis of the  time series to capture the temporal features of the original hidden sequence $H_{1:T}^{Real}$ and generate the temporal feature vector $H_{emb}^{Real}$. The generator, through adversarial training, is able to map the input random sequence $Z$ to the latent space where the temporal feature vector resides, generating $H_{emb}^{Fake}$. By extracting the process of time series feature extraction from the generator of the generative adversarial network and placing it on the original hidden sequence, we ensure the effectiveness of the temporal feature extraction process. Simultaneously, it avoids performing temporal feature extraction modules on immature time series data generated by the generator.

% 时序特征提取器的详细结构如图2c所示。对于多维时间序列而言，不同通道的时间序列具有一定通道独立性的同时又相互影响 47，因此，我们首先将输入的多为时间序列视作多个单变量时间序列，进行通道独立的时间序列建模；随后，为了捕获跨通道的时间序列关联性，对进行过单变量时序建模的时间序列进行拼接还原为多维时间序列，并对同一时间步不同信息通道之间的关联性进行捕获。
The detailed structure of the temporal feature extractor is illustrated in Figure\ref{fig:extractor}(c. For multivariate time series, the time series of different channels exhibit a certain degree of channel independence while also influencing each other\cite{47}. Therefore, we first treat the input multivariate time series $H_{1:T}^{Real}$ as multiple univariate time series and perform channel-independent temporal modeling. Subsequently, to capture the cross-channel correlations, the univariate-modeled time series are concatenated and restored as multivariate time series, enabling the capture of correlations between different information channels at the same time step.

% 在通道独立建模中，由于时间序列其时序特征往往反应在其动态变化中，而单一时间步上的取值并没有丰富的时间语义，因此，对于输入的单变量时间序列，我们选择合适大小的滑动窗口对时间序列进行无重叠的划分，将其切割为时间片段，保留时间动态的同时也保留了时间序列的局部依赖性。随后，我们使用多头自注意力机制进行特征增强，捕获时间序列的全局依赖性以及周期性。
In channel-independent modeling, since the temporal features of a time series are often reflected in its dynamic changes, the value at a single time step does not carry rich temporal semantics\cite{11,13}. Therefore, for the input univariate time series, we select an appropriately sized sliding window to partition the time series into non-overlapping segments, preserving the temporal dynamics while maintaining the local dependencies of the series. Subsequently, we employ a multi-head self-attention mechanism(MSA) to enhance the features, capturing the global dependencies and periodicity of the time series.

\begin{equation}
 H_{:,d}^{time}=MSA\left ( PATCH \left( H_{:,d}\right ) + E_{:,d}^{pos} \right), 
\end{equation}

% 其中，patch表示使用滑动窗口进行的切割操作，自注意力机制将不同位置的推理距离缩短为1，虽然有效捕获了输入之间的全局依赖性，但忽略了位置信息，这在时间维度上是及其重要的，因此我们为不同的patch加入了位置编码。
In the above equation, $H_{:,d}$ represents the value of the hidden sequence on data dimension $d$, $PATCH$ is the sliding window partitioning operation. While the self-attention mechanism effectively reduces the reasoning distance between different positions to $1$, capturing the global dependencies of the input, it neglects positional information, which is crucial in the temporal dimension. To address this, we incorporate positional encoding $E_{:,d}^{pos}$ into each patch.

% 在跨通道建模中，我们将不同通道的时间序列进行合并，但仍保持了patch划分状态，同样利用自注意力机制来捕获不同通道在同一时间位置上patch的依赖性，不同的是，由于不同通道之间并没有显式的位置关系，因此，我们并没有添加位置编码，以此来使其更符合时间序列建模。

In cross-channel modeling, we merge the time series from different channels while maintaining the patch segmentation. Similarly, we use a self-attention mechanism to capture the dependencies of patches across different channels at the same temporal position. However, unlike in the temporal dimension, there is no explicit positional relationship between channels. Thus, we omit positional encoding, ensuring the modeling aligns better with the characteristics of time series data.
\begin{equation}
 H_{t,:}^{dim}=MSA\left ( H_{t,:}^{time} \right) .
\end{equation}

% 注意力机制虽然在捕获全局依赖性上表现优越，但其本质上还是建立在离散建模的基础之上。因此，最后我们采用深度GRU网络进行的时间序列建模以及舒徐特征的提取。
Although the attention mechanism excels in capturing global dependencies, it fundamentally relies on discrete modeling. Therefore, we employ a deep GRU network for temporal modeling and the extraction of temporal features.
\begin{equation}
 H_{emb}^{Real}=GRU(H^{dim}_{:,:}).
\end{equation}

The detailed structure of the Generator\textsubscript{1} is illustrated in Figure\ref{fig:extractor}(a. We adopted a trend-first modeling approach followed by detail enhancement to imbue random sequences $Z$ with temporal dependencies. First, we utilized an average pooling window with a stride of 1 to extract relatively smooth trend information. The trend information was then used as the query, while the detailed information served as the key and value. A cross-attention mechanism was employed to construct the synthetic time series, which was further processed by a deep GRU network for temporal feature extraction.
\begin{equation}
  H_{emb}^{Fake}=GRU( CA (TF(AvgPool(Z)), TF(Z), TF(Z)),
\end{equation}
\begin{equation}
  CA(Q, K, V) = Softmax(Q\times K^T)\times V,
\end{equation}
Here, $TF$ refers to the vanilla Transformer encoder. Discriminator\textsubscript{1} distinguishes between the original temporal feature vectors $H_{emb}^{Real}$ and the synthetic temporal feature vectors $H_{emb}^{Fake}$ to enhance the synthesis quality of Generator\textsubscript{1} and the discriminative ability of Discriminator\textsubscript{1}.
\begin{equation}
  y_1=Discriminator_1 (H_{emb}^{Real},H_{emb}^{Fake} )
\end{equation}

% Generator\textsubscript{1} accepts a random sequence $Z$ and generates the synthetic time series feature vector $H_{emb}^{Fake}$ through time series modeling. The reason for choosing the time series modeling process is solely to ensure that both outputs have similar formats, while ensuring that the inputs are equally informative. Discriminator\textsubscript{1} distinguishes between the original temporal feature vectors $H_{emb}^{Real}$ and the synthetic temporal feature vectors $H_{emb}^{Fake}$ to enhance the synthesis quality of Generator\textsubscript{1} and the discriminative ability of Discriminator\textsubscript{1}.

% \begin{equation}
%   H_{emb}^{Fake}=Generator_1 (Z)
% \end{equation}
% \begin{equation}
%   y_1=Discriminator_1 (H_{emb}^{Real},H_{emb}^{Fake} )
% \end{equation}

% Through the aforementioned approach, the temporal feature extractor is capable of capturing the temporal features of the original time series. Meanwhile, Generator\textsubscript{1}, through adversarial training, is able to generate synthetic temporal feature vectors that aligned with the original feature vectors in the distribution space. This process not only ensures the effectiveness of the temporal feature extraction process but also enables unsupervised adversarial learning to acquire knowledge about the features of the original sequences.

\subsection{Sequence Reconstructor}

% 生成器2同时接受来生生成器1以及特征提取器产生的时序特征向量，采用迭代子生成方式来重建目标序列，同时利用teacher forceing方式来重建真实序列，保证能够充分学习到真实的时序依赖。
Sequence Reconstructor mainly consists of Generator\textsubscript{2} and Discriminator\textsubscript{2}. Generator\textsubscript{2} , also known as the Feature Reconstructor, is illustrated in Figure\ref{fig:extractor}(b. It receives the time series feature vectors outputted by Generator\textsubscript{1} and the Temporal feature extractor, adoptting an iterative autoregressive generation approach to reconstruct the target time series step by step. Additionally, we utilized the teacher forcing approach to reconstruct the real sequence, ensuring the model effectively learns the authentic temporal dependencies. Discriminator\textsubscript{2} receives three types of inputs: $ \widehat{H}_{1:T}^{Real }$ and $ \widehat{H}_{1:T}^{Fake }$ generated by the feature reconstructor from $H_{emb}^{Real}$ and $H_{emb}^{Fake}$ respectively, and $H_{1:T}^{Real}$ generated by the encoder from $X_{1:T}$. Adversarial training enhances the generative capability of Generator\textsubscript{2} and the discriminative ability of Discriminator\textsubscript{2}.

\begin{equation}
  \widehat{H}_{1:T}^{Fake}=Generator_2 (H_{emb}^{Fake} ),
\end{equation}
\begin{equation}
  \widehat{H}_{1:T}^{Real}=Generator_2 (H_{emb}^{Real} ),
\end{equation}
\begin{equation}
  y_2=Discriminator_2 (H_{1:T}^{Real},H_{1:T}^{Real},H_{1:T}^{Fake} ).
\end{equation}

\section{Training Procedure}
During the training process of the model, we utilized two types of loss functions. Firstly, we employed the Mean Squared Error (MSE) loss function $L_{mse}$ as the supervised loss function for time series reconstruction error. For the GAN modules in the model, we utilized the standard GAN loss $L_{gan}$. We trained DLGAN according to the following procedure:

\begin{enumerate}
    \item Firstly, we pre-trained the Sequence Autoencoder. We trained the encoder-decoder pair based on the reconstruction loss $L_R^{AE}$ of the Autoencoder.
    \begin{equation}
        L_R^{AE}=L_{mse} (X_{1:T},\widehat{X}_{1:T}^{Real} ).
    \end{equation}
    \item Next, to ensure the effectiveness of the Temporal feature extractor in extracting temporal feature vectors, we pre-trained the Temporal feature extractor and Feature Reconstructor. The loss error was the reconstruction loss $L_R^H$ of the embedding sequences $H_{1:T}^{Real}$.
    \begin{equation}
        L_R^H=L_{mse} (H_{1:T}^{Real},\widehat{H} _{1:T}^{Real} ).
    \end{equation}
    \item After the pre-training steps mentioned above, we proceeded with joint training of all modules. Each iteration mainly involved three processes: training the Generators, training the Autoencoder, and training the Discriminators. Firstly, we trained Generator\textsubscript{1} and Generator\textsubscript{2} using the standard GAN loss. To ensure that the training effectiveness does not solely rely on unsupervised adversarial training, we incorporated a supervised reconstruction loss $L_R^H$ for the original hidden sequence $H_{1:T}^{Real}$. This ensures that the Generators can learn the feature distribution of original sequences. Therefore, the Generator loss function $L_G$ is specifically defined as follows:
    \begin{equation}
        L_G=L_{gan} (y_1 )+L_{gan} (y_2 )+L_{mse} (H_{1:T}^{Real},\widehat{H}_{1:T}^{Real} ).
    \end{equation}
    Next, we further fine-tuned the Encoder and Decoder in the Sequence Autoencoder using the reconstruction losses$ L_R^{AE}$ and $L_R^H$ from steps 1 and 2. This ensures that the Encoder can generate temporal embedding structures suitable for subsequent module learning. Finally, we trained the Discriminators using the standard GAN loss. Therefore, the discriminator loss function $L_D$ can be represented as follows:
    \begin{equation}
        L_D=L_{gan} (y_1 )+L_{gan} (y_2 ).
    \end{equation}
\end{enumerate}
Through the aforementioned training process, random sequences can be synthesized into the target time series dataset via Generator\textsubscript{1}, Generator\textsubscript{2}, and the Decoder.

\begin{figure*}[hbtp]
  \centering
  \includegraphics[width=0.9\textwidth]{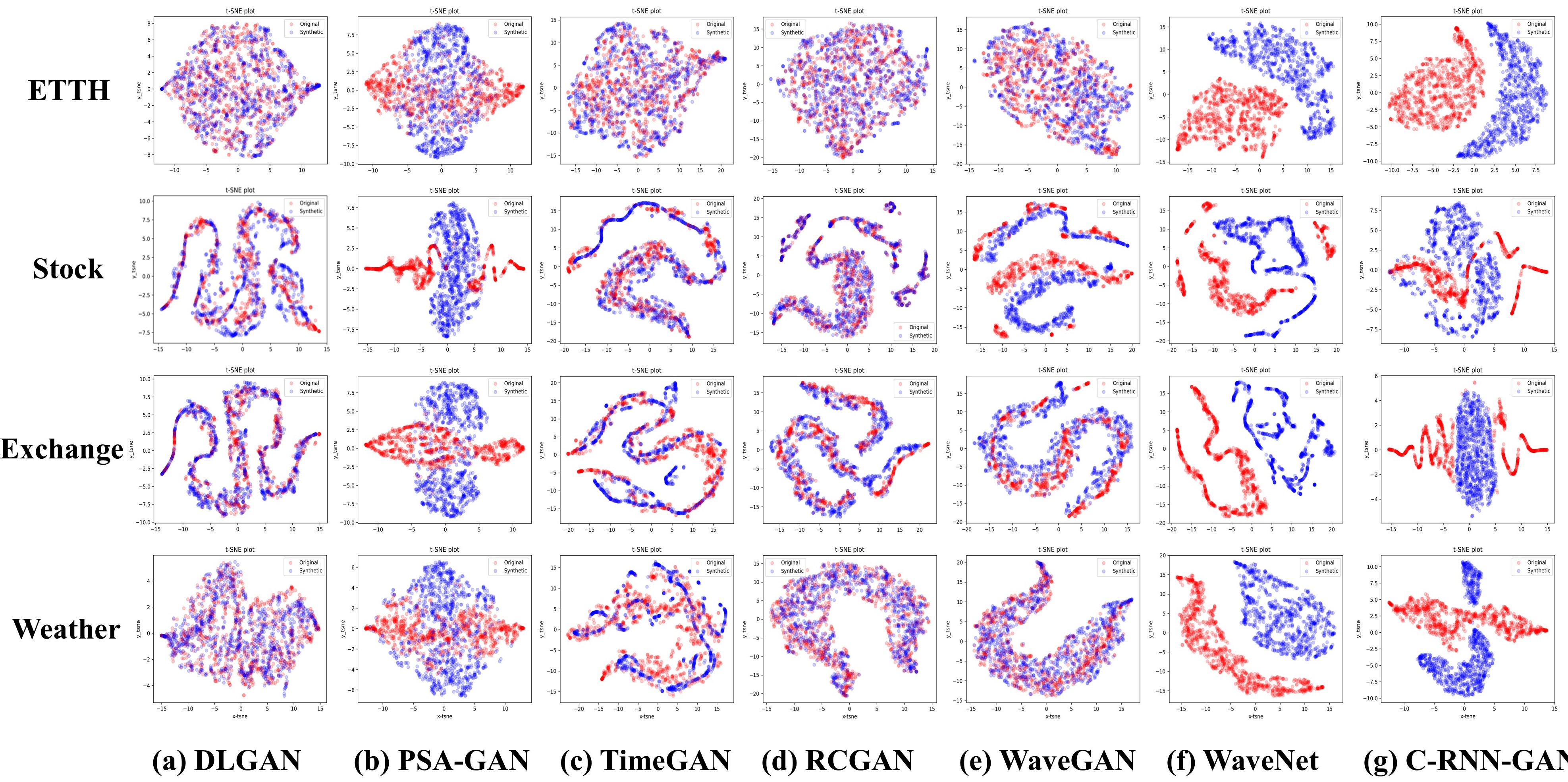}
  \caption{t-SNE visualization. Red denotes original data, and blue denotes synthetic data.}
  \label{fig:result}
\end{figure*}

\section{Experiments}
\subsection{Experiment Setup}
\textbf{Dataset.} We conducted experiments on four publicly available time series datasets: \textbf{1)ETTH} \cite{14}(Electricity Transformer Temperature-hourly), \textbf{2)Stock} \cite{9} (Google Stock data-daily), \textbf{3)Exchange} \cite{16} Excahnge rate data-daily, \textbf{4)weather} \cite{15} Weather data-minutely. These datasets have been widely used in various time series analysis tasks.

\textbf{Baseline.} We selected the following six sequence generation models as baselines: \textbf{1)PSA-GAN} \cite{10}, \textbf{2)TimeGAN} \cite{9}, \textbf{3)RCGAN} \cite{39}, \textbf{4)WaveGAN} \cite{17},\textbf{5)WaveNet} \cite{40},\textbf{6)C-RNN-GAN} \cite{8}

\textbf{Evaluation Metrics.} This paper adopts the synthetic data evaluation metrics from the study \cite{9},  which mainly include the following three evaluation metrics: 
\begin{enumerate}
\item  \textbf{Visualization}: Using the t-SNE\cite{18}, multivariate time series can be mapped to a two-dimensional space for visualizing the distribution of real and synthetic data. By comparing the overlap between the two distributions, the quality of the synthetic data can be assessed.
\item  \textbf{Discriminative Score}: Used to measure the similarity between the original and synthetic data. A posterior classifier based GRU is employed to differentiate between real and synthetic data. The discrimination score is defined as the absolute difference between the classification accuracy and 0.5.  A lower discrimination score indicates higher quality of the synthetic data.
\item  \textbf{Prediction Score}: The main idea is that synthetic data should exhibit the same performance as the original data when facing the same prediction task. Train a GRU-based predictor on the synthetic dataset and test it on the real dataset. A smaller prediction error indicates higher quality of the synthetic data.
\end{enumerate}

\subsection{Experiment Results}
\textbf{Visualization.} The experimental results are depicted in Figure \ref{fig:result}. On all four datasets, the synthetic datasets generated by C-RNN-GAN and WaveNet exhibit a significant discrepancy compared to the real data. PSA-GAN, due to its progressive growing mode, may lose some temporal patterns, resulting in a relatively high overlap between its synthetic datasets and the original datasets in the data centre, but it often fails to fully cover the original datasets. DLGAN, TimeGAN, RCGAN and WaveGAN achieve relatively good synthesis results on multiple datasets. However, WaveGAN performs poorly on datasets with insufficient data volume like Stock, while TimeGAN's synthesis performance deteriorates when dealing with high-dimensional datasets like Weather. RCGAN benefits from benefiting from conditional generation, achieves relatively satisfactory synthetic results. DLGAN  achieves satisfactory synthetic results on all datasets.

\textbf{Discriminative Score and Prediction Score.} The experimental results are presented in Table \ref{tab:result1}. PSA-GAN uses the attention mechanism to capture global dependencies but often overlooks local sequence dependencies. Additionally, the progressive growing mode from coarse-grained modeling to fine-grained modeling may lead to the loss of some multi-period patterns in time series, resulting in less than ideal results. WaveNet primarily learns information from the original dataset, but it overlooks the ability to map random input sequences to the target space, which results in mediocre overall performance. WaveGAN uses transposed convolutions to generate sequence data, including a process where the sequence gradually grows.  It maintains consistent data information throughout, leading to better predictive performance compared to PSA-GAN. However, due to the lack of continuity modeling, its discrimination score is not ideal. C-RNN-GAN reconstructs GAN framework using an RNNs, ensuring basic continuity in the generated sequences. However, due to its relatively low model complexity, its performance tends to be unstable. RCGAN benefits from conditional generation and generally outperforms the aforementioned methods, but it still does not consider temporal features inherent in time series. TimeGAN incorporates temporal dynamics support during the generation process, achieving better experimental performance compared to the aforementioned methods. However, its overall performance may deteriorate when handling high-dimensional datasets like Weather. DLGAN fully considers temporal features of original dataset and intentionally injects temporal dependencies into the synthesized sequences, achieving  overall better experimental results than the baselines.

\begin{table*}[h]
\centering
  \caption{Results on Multiple Time-Series Dataset (Bold indicates best performance)}
  \label{tab:result1}
  \begin{tabular}{c|c|c|c|c|c}
    \toprule
    Metric&Methed&ETTH&Stock&Exchange&Weather\\
    \midrule
    \multirow{7}{*}{\makecell{Discriminative\\\\Score}} &C-RNN-GAN&0.497&0.399&0.289&0.487\\
    &WaveNet&0.332&0.232&0.394&0.498\\
    &WaveGAN&0.252&0.244&0.293&0.425\\
    &RCGAN&0.125&0.196&0.317&0.493\\
    &TimeGAN&0.106&0.102&0.209&0.412\\
    &PSA-GAN&0.497&0.469&0.499&0.499\\
    &DLGAN&\textbf{0.079}&\textbf{0.090}&\textbf{0.104}&\textbf{0.173}\\

    \midrule
    \multirow{7}{*}{\makecell{Predictive\\\\Score}} &C-RNN-GAN&0.131&0.038&0.179&0.007\\
    &WaveNet&0.347&0.042&0.252&0.527\\
    &WaveGAN&0.156&0.040&0.068&0.019\\
    &RCGAN&0.146&0.040&0.061&0.006\\
    &TimeGAN&0.132&0.038&0.051&\textbf{0.002}\\
    &PSA-GAN&0.253&0.061&0.331&0.484\\
    &DLGAN&\textbf{0.127}&\textbf{0.037}&\textbf{0.048}&0.003\\
   
  \bottomrule
\end{tabular}
\end{table*}

\begin{table*}[h]
    \centering
  \caption{Results of Ablation Experiments (Bold indicates best performance)}
  \label{tab:result2}
  \begin{tabular}{c|c|c|c|c|c}
    \toprule
    Metric&Methed&ETTH&Stock&Exchange&Weather\\
    \midrule
    \multirow{4}{*}{\makecell{Discriminative\\\\Score}} &w/o Temporal Feature Extractor&0.086&0.152&0.155&0.267\\
    &w/o TimeSeries Reconstructor&0.134&0.169&0.163&0.324\\
    &w/o All&0.095&0.181&0.127&0.177\\
    &DLGAN&\textbf{0.079}&\textbf{0.090}&\textbf{0.104}&\textbf{0.173}\\

    \midrule
    \multirow{4}{*}{\makecell{Predictive\\\\Score}} &w/o Temporal Feature Extractor&0.133&0.039&0.066&0.003\\
    &w/o TimeSeries Reconstructor&0.177&0.038&0.092&\textbf{0.002}\\
    &w/o All&0.130&0.040&0.054&0.003\\
    &DLGAN&\textbf{0.127}&\textbf{0.037}&\textbf{0.048}&0.003\\
  \bottomrule
\end{tabular}
\end{table*}

\subsection{Ablation Experiments}
The main purpose of DLGAN is to fully consider the temporal features of time series while learning the mapping relationship from random sequences to target sequences, and generate synthetic time series that genuinely exhibit temporal dependencies. Its gains stem mainly from two aspects: 1) the Temporal feature extractor, which captures temporal features effectively through multi-patch partitioning and a two-stage attention mechanism, thereby enhancing the temporal dependencies of input sequences; 2) time series reconstructor, which uses an iterative self-generation method to reconstruct the time series, explicitly ensuring the temporal dependencies of the synthesized sequences through supervised learning based on original dataset. Therefore, we analyze the effectiveness of these two components by removing them from the model. The experimental results are shown in Table \ref{tab:result2}. \textbf{w/o Feature Extractor} indicates that we remove the Feature Extractor, and only retained the most basic sequence modeling process to extract time feature vectors. \textbf{w/o TimeSeries Reconstructor} indicates that we remove the feature vector extraction and self-generation processes for time series reconstruction. Instead, we used the sequence data generated by the Temporal feature extractor and Generator\textsubscript{1} through temporal modeling as input for Generator\textsubscript{2}. \textbf{w/o ALL} indicates the simultaneous removal of both components, where the Temporal feature extractor retains only the simplest sequence modeling process and Generator\textsubscript{2} takes sequence data as input.

The results of the ablation experiment are presented in Table \ref{tab:result2}. From the experimental results, it is evident that the complete DLGAN model achieved the best results on all four datasets. However, on some datasets, the \textbf{w/o All} were superior to those obtained by solely removing single mechanisms. The analysis of the structure of the model shows that in order to ensure the temporal dependence of the generated sequences, the temporal feature extraction and reconstruction processes we employed condensed the information of input sequences into temporal feature vectors. While this approach guarantees the temporal dependency of reconstructed sequences, it inevitably results in the loss of some information in the input sequences. Nevertheless, Our model still achieve results that far exceed those of the baselines, demonstrating that even the simplest sequence modeling process can still improve the quality of synthesised time series data through supervised learning based on the original data. This further demonstrates the effectiveness of temporal feature extraction and time series reconstruction mechanisms in ensuring the temporal dependency of the synthesized sequences.

\section{Conclusion}
In this paper, we propose a simple but effective generative model called DLGAN, which introduces a stacked GAN structure to decompose the time series synthesis process, effectively addressing the issue of insufficient extraction of temporal features in generation processes while learning the mapping relationship from random sequences to target sequences, and explicitly incorporates the temporal dependencies of the original dataset into the synthetic time series through a supervised learning process. Extensive experiments demonstrate that DLGAN outperforms the state-of-the-art methods in time series synthesis tasks, and ablation studies validate the effectiveness of each module in the model. In the future, we will further investigate privacy protection for synthesized time series data and the integrity of the synthesized data distribution patterns to generate higher-quality time series data.


\end{sloppypar}

\begin{thebibliography}{100}

\bibitem{26}
Moustafa Alzantot, Supriyo Chakraborty, and Mani Srivastava.
\newblock Sensegen: A deep learning architecture for synthetic sensor data generation.
\newblock In {\em 2017 IEEE International Conference on Pervasive Computing and Communications Workshops (PerCom Workshops)}, pages 188--193. IEEE, 2017.


\bibitem{35}
Max Baak, Simon Brugman, Ilan~Fridman Rojas, Lorraine Dalmeida, Ralph~EQ Urlus, and Jean-Baptiste Oger.
\newblock Synthsonic: Fast, probabilistic modeling and synthesis of tabular data.
\newblock In {\em International Conference on Artificial Intelligence and Statistics}, pages 4747--4763. PMLR, 2022.



\bibitem{23}
Kuntai Cai, Xiaoyu Lei, Jianxin Wei, and Xiaokui Xiao.
\newblock Data synthesis via differentially private markov random fields.
\newblock {\em Proceedings of the VLDB Endowment}, 14(11):2190--2202, 2021.


\bibitem{31}
Dingfan Chen, Ning Yu, Yang Zhang, and Mario Fritz.
\newblock Gan-leaks: A taxonomy of membership inference attacks against generative models.
\newblock In {\em Proceedings of the 2020 ACM SIGSAC conference on computer and communications security}, pages 343--362, 2020.


\bibitem{30}
Xue Chen, Cheng Wang, Qing Yang, Teng Hu, and Changjun Jiang.
\newblock Locally differentially private high-dimensional data synthesis.
\newblock {\em Science China Information Sciences}, 66(1):112101, 2023.


\bibitem{5}
Junyoung Chung, Caglar Gulcehre, Kyunghyun Cho, and Yoshua Bengio.
\newblock Empirical evaluation of gated recurrent neural networks on sequence modeling.
\newblock In {\em NIPS 2014 Workshop on Deep Learning, December 2014}, 2014.

\bibitem{43}
Abhyuday Desai, Cynthia Freeman, Zuhui Wang, and Ian Beaver.
\newblock Timevae: A variational auto-encoder for multivariate time series generation.
\newblock {\em arXiv preprint arXiv:2111.08095}, 2021.

\bibitem{17}
Chris Donahue, Julian McAuley, and Miller Puckette.
\newblock Adversarial audio synthesis.
\newblock In {\em International Conference on Learning Representations}, 2019.


\bibitem{36}
Yuntao Du, Jindong Wang, Wenjie Feng, Sinno Pan, Tao Qin, Renjun Xu, and Chongjun Wang.
\newblock Adarnn: Adaptive learning and forecasting of time series.
\newblock In {\em Proceedings of the 30th ACM international conference on information \& knowledge management}, pages 402--411, 2021.

\bibitem{39}
Crist{\'o}bal Esteban, Stephanie~L Hyland, and Gunnar R{\"a}tsch.
\newblock Real-valued (medical) time series generation with recurrent conditional gans.
\newblock {\em arXiv preprint arXiv:1706.02633}, 2017.

\bibitem{25}
Ju~Fan, Junyou Chen, Tongyu Liu, Yuwei Shen, Guoliang Li, and Xiaoyong Du.
\newblock Relational data synthesis using generative adversarial networks.
\newblock {\em Proceedings of the VLDB Endowment}, 13(12):1962--1975, 2020.


\bibitem{3}
Ian Goodfellow, Jean Pouget-Abadie, Mehdi Mirza, Bing Xu, David Warde-Farley, Sherjil Ozair, Aaron Courville, and Yoshua Bengio.
\newblock Generative adversarial nets.
\newblock {\em Advances in neural information processing systems}, 27, 2014.


\bibitem{47}
Lu~Han, Han-Jia Ye, and De-Chuan Zhan.
\newblock The capacity and robustness trade-off: Revisiting the channel independent strategy for multivariate time series forecasting.
\newblock {\em IEEE Transactions on Knowledge and Data Engineering}, 2024.


\bibitem{42}
Jonathan Ho, Ajay Jain, and Pieter Abbeel.
\newblock Denoising diffusion probabilistic models.
\newblock {\em Advances in neural information processing systems}, 33:6840--6851, 2020.

\bibitem{4}
Sepp Hochreiter and J{\"u}rgen Schmidhuber.
\newblock Long short-term memory.
\newblock {\em Neural computation}, 9(8):1735--1780, 1997.


\bibitem{10}
Paul Jeha, Michael Bohlke-Schneider, Pedro Mercado, Shubham Kapoor, Rajbir~Singh Nirwan, Valentin Flunkert, Jan Gasthaus, and Tim Januschowski.
\newblock Psa-gan: Progressive self attention gans for synthetic time series.
\newblock In {\em International Conference on Learning Representations}, 2022.


\bibitem{21}
Tero Karras, Timo Aila, Samuli Laine, and Jaakko Lehtinen.
\newblock Progressive growing of gans for improved quality, stability, and variation.
\newblock In {\em International Conference on Learning Representations}, 2018.


\bibitem{41}
Diederik~P Kingma and Max Welling.
\newblock Auto-encoding variational bayes.
\newblock {\em arXiv preprint arXiv:1312.6114}, 2013.


\bibitem{16}
Guokun Lai, Wei-Cheng Chang, Yiming Yang, and Hanxiao Liu.
\newblock Modeling long-and short-term temporal patterns with deep neural networks.
\newblock In {\em The 41st international ACM SIGIR conference on research \& development in information retrieval}, pages 95--104, 2018.



\bibitem{7}
Jaehoon Lee, Jihyeon Hyeong, Jinsung Jeon, Noseong Park, and Jihoon Cho.
\newblock Invertible tabular gans: Killing two birds with one stone for tabular data synthesis.
\newblock {\em Advances in Neural Information Processing Systems}, 34:4263--4273, 2021.


\bibitem{44}
Hongming Li, Shujian Yu, and Jose Principe.
\newblock Causal recurrent variational autoencoder for medical time series generation.
\newblock In {\em Proceedings of the AAAI conference on artificial intelligence}, volume~37, pages 8562--8570, 2023.



\bibitem{27}
Zachary~C Lipton, John Berkowitz, and Charles Elkan.
\newblock A critical review of recurrent neural networks for sequence learning.
\newblock {\em arXiv preprint arXiv:1506.00019}, 2015.


\bibitem{8}
Olof Mogren.
\newblock C-rnn-gan: Continuous recurrent neural networks with adversarial training.
\newblock {\em arXiv preprint arXiv:1611.09904}, 2016.


\bibitem{11}
Yuqi Nie, Nam~H Nguyen, Phanwadee Sinthong, and Jayant Kalagnanam.
\newblock A time series is worth 64 words: Long-term forecasting with transformers.
\newblock In {\em The Eleventh International Conference on Learning Representations}, 2023.


\bibitem{6}
Noseong Park, Mahmoud Mohammadi, Kshitij Gorde, Sushil Jajodia, Hongkyu Park, and Youngmin Kim.
\newblock Data synthesis based on generative adversarial networks.
\newblock {\em Proceedings of the VLDB Endowment}, 11(10), 2018.


\bibitem{20}
Alec Radford, Luke Metz, and Soumith Chintala.
\newblock Unsupervised representation learning with deep convolutional generative adversarial networks.
\newblock {\em arXiv preprint arXiv:1511.06434}, 2015.



\bibitem{32}
Shulan Ruan, Yong Zhang, Kun Zhang, Yanbo Fan, Fan Tang, Qi~Liu, and Enhong Chen.
\newblock Dae-gan: Dynamic aspect-aware gan for text-to-image synthesis.
\newblock In {\em Proceedings of the IEEE/CVF International Conference on Computer Vision}, pages 13960--13969, 2021.


\bibitem{19}
Mohammad~Amin Shabani, Amir~H Abdi, Lili Meng, and Tristan Sylvain.
\newblock Scaleformer: Iterative multi-scale refining transformers for time series forecasting.
\newblock In {\em International Conference on Learning Representations}, 2023.



\bibitem{46}
Lifeng Shen, Weiyu Chen, and James Kwok.
\newblock Multi-resolution diffusion models for time series forecasting.
\newblock In {\em The Twelfth International Conference on Learning Representations}, 2024.


\bibitem{28}
Shun Takagi, Tsubasa Takahashi, Yang Cao, and Masatoshi Yoshikawa.
\newblock P3gm: Private high-dimensional data release via privacy preserving phased generative model.
\newblock In {\em 2021 IEEE 37th International Conference on Data Engineering (ICDE)}, pages 169--180. IEEE, 2021.


\bibitem{40}
Aäron {van den Oord}, Sander Dieleman, Heiga Zen, Karen Simonyan, Oriol Vinyals, Alex Graves, Nal Kalchbrenner, Andrew Senior, and Koray Kavukcuoglu.
\newblock {WaveNet: A Generative Model for Raw Audio}.
\newblock In {\em Proc. 9th ISCA Workshop on Speech Synthesis Workshop (SSW 9)}, page 125, 2016.


\bibitem{18}
Laurens Van~der Maaten and Geoffrey Hinton.
\newblock Visualizing data using t-sne.
\newblock {\em Journal of machine learning research}, 9(11), 2008.



\bibitem{22}
Zhendong Wang, Huangjie Zheng, Pengcheng He, Weizhu Chen, and Mingyuan Zhou.
\newblock Diffusion-gan: Training gans with diffusion.
\newblock In {\em International Conference on Learning Representations}, 2023.


\bibitem{34}
Qingsong Wen, Tian Zhou, Chaoli Zhang, Weiqi Chen, Ziqing Ma, Junchi Yan, and Liang Sun.
\newblock Transformers in time series: a survey.
\newblock In {\em Proceedings of the Thirty-Second International Joint Conference on Artificial Intelligence}, pages 6778--6786, 2023.


\bibitem{15}
Haixu Wu, Jiehui Xu, Jianmin Wang, and Mingsheng Long.
\newblock Autoformer: Decomposition transformers with auto-correlation for long-term series forecasting.
\newblock {\em Advances in neural information processing systems}, 34:22419--22430, 2021.



\bibitem{29}
Yang Yang, Ke~Mu, and Robert~H Deng.
\newblock Lightweight privacy-preserving gan framework for model training and image synthesis.
\newblock {\em IEEE Transactions on Information Forensics and Security}, 17:1083--1098, 2022.

\bibitem{9}
Jinsung Yoon, Daniel Jarrett, and Mihaela Van~der Schaar.
\newblock Time-series generative adversarial networks.
\newblock {\em Advances in neural information processing systems}, 32, 2019.

\bibitem{37}
Xiaoyu You, Mi~Zhang, Daizong Ding, Fuli Feng, and Yuanmin Huang.
\newblock Learning to learn the future: Modeling concept drifts in time series prediction.
\newblock In {\em Proceedings of the 30th ACM International Conference on Information \& Knowledge Management}, pages 2434--2443, 2021.

\bibitem{38}
Chengqing Yu, Fei Wang, Zezhi Shao, Tao Sun, Lin Wu, and Yongjun Xu.
\newblock Dsformer: A double sampling transformer for multivariate time series long-term prediction.
\newblock In {\em Proceedings of the 32nd ACM International Conference on Information and Knowledge Management}, pages 3062--3072, 2023.

\bibitem{33}
Ning Yu, Larry~S Davis, and Mario Fritz.
\newblock Attributing fake images to gans: Learning and analyzing gan fingerprints.
\newblock In {\em Proceedings of the IEEE/CVF international conference on computer vision}, pages 7556--7566, 2019.

\bibitem{45}
Xinyu Yuan and Yan Qiao.
\newblock Diffusion-{TS}: Interpretable diffusion for general time series generation.
\newblock In {\em The Twelfth International Conference on Learning Representations}, 2024.

\bibitem{24}
Chongsheng Zhang, Yaxin Hou, Ke~Chen, Shuang Cao, Gaojuan Fan, and Ji~Liu.
\newblock Quality-aware self-training on differentiable synthesis of rare relational data.
\newblock In {\em Proceedings of the AAAI Conference on Artificial Intelligence}, volume~37, pages 6602--6611, 2023.

\bibitem{2}
Jun Zhang, Graham Cormode, Cecilia~M Procopiuc, Divesh Srivastava, and Xiaokui Xiao.
\newblock Privbayes: Private data release via bayesian networks.
\newblock {\em ACM Transactions on Database Systems (TODS)}, 42(4):1--41, 2017.

\bibitem{13}
Yunhao Zhang and Junchi Yan.
\newblock Crossformer: Transformer utilizing cross-dimension dependency for multivariate time series forecasting.
\newblock In {\em The eleventh international conference on learning representations}, 2022.


\bibitem{14}
Haoyi Zhou, Shanghang Zhang, Jieqi Peng, Shuai Zhang, Jianxin Li, Hui Xiong, and Wancai Zhang.
\newblock Informer: Beyond efficient transformer for long sequence time-series forecasting.
\newblock In {\em Proceedings of the AAAI conference on artificial intelligence}, volume~35, pages 11106--11115, 2021.





\end{thebibliography}
\end{document}